\pdfoutput=1

\documentclass[11pt]{article}

\usepackage{acl}

\usepackage{times}
\usepackage{latexsym}

\usepackage[T1]{fontenc}

\usepackage[utf8]{inputenc}

\usepackage{microtype}

\usepackage{graphicx,caption}
\usepackage{url}
\usepackage{amsmath,amssymb}
\usepackage{multirow,booktabs}
\usepackage{makecell}
\usepackage{CJKutf8}

\title{Sememe Prediction for BabelNet Synsets using Multilingual \\ and Multimodal Information}

\author{
Fanchao Qi$^{1}$\thanks{\ \ Equal contribution}\hspace{0.3em}, Chuancheng Lv$^{2,5*}$, Zhiyuan Liu$^{1,3,4}$, Xiaojun Meng$^{6}$, \\
{\bf Maosong Sun$^{1,3,4}$\thanks{\ \  Corresponding author. Email: sms@tsinghua.edu.cn}\hspace{0.3em},
 Hai-Tao Zheng$^{2,5}$
}\\
$^{1}$Dept. of Comp. Sci. \& Tech., Institute for AI, Tsinghua University, Beijing, China\\
Beijing National Research Center for Information Science and Technology\\
$^{2}$Shenzhen International Graduate School, Tsinghua University, China \\
$^{3}$Institute Guo Qiang, Tsinghua University, Beijing, China \\
$^{4}$International Innovation Center of Tsinghua University, Shanghai, China \\
$^{5}$Peng Cheng Laboratory \quad $^{6}$Huawei Noah's Ark Lab\\
{\tt  \{qfc17,lvcc21\}@mails.tsinghua.edu.cn}
}

\begin{document}
\maketitle

\begin{abstract}		
In linguistics, a sememe is defined as the minimum semantic unit of languages.
Sememe knowledge bases (KBs), which are built by manually annotating words with sememes, have been successfully applied to various NLP tasks.
However, existing sememe KBs only cover a few languages, which hinders the wide utilization of sememes.
To address this issue, the task of sememe prediction for BabelNet synsets (SPBS) is presented, aiming to build a multilingual sememe KB based on BabelNet, a multilingual encyclopedia dictionary.
By automatically predicting sememes for a BabelNet synset, the words in many languages in the synset would obtain sememe annotations simultaneously.
However, previous SPBS methods have not taken full advantage of the abundant information in BabelNet.
In this paper, we utilize the multilingual synonyms, multilingual glosses and images in BabelNet for SPBS.
We design a multimodal information fusion model to encode and combine this information for sememe prediction.
Experimental results show the substantial outperformance of our model over previous methods (about 10 MAP and F1 scores).
All the code and data of this paper can be obtained at \url{https://github.com/thunlp/MSGI}.

\end{abstract}

\section{Introduction}
\begin{CJK}{UTF8}{gkai}

A word is the smallest unit of language that can stand on its own \citep{o1997contemporary}, but its meaning can be further divided into smaller components.
In linguistics, a \textit{sememe} is defined as the minimum semantic unit \citep{bloomfield1926set}.
It is believed by some linguists that the meanings of all the words in any language can be decomposed of a limited set of language-independent sememes, which is equated with the idea of semantic primitives \citep{wierzbicka1996semantics}.

\begin{figure}[t]
\centering
	\includegraphics[width=\linewidth]{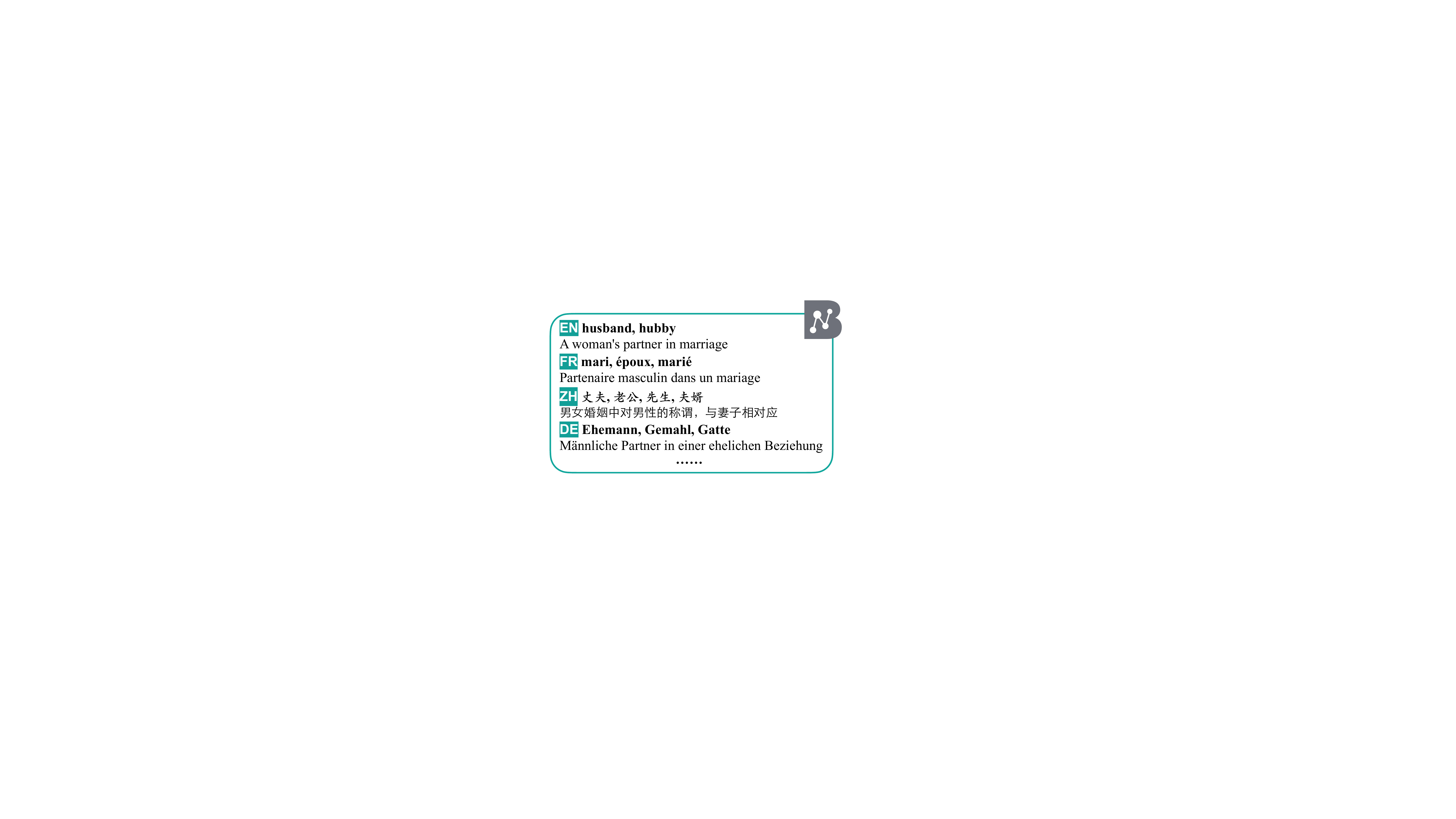}
	\caption{Sememe annotations of the English word ``husband'' in HowNet. For succinctness, we only show the English notations of sememes, although sememes have both English and Chinese notations in HowNet, e.g., \texttt{family|家庭}.}
	\label{fig:hownet}
\end{figure}

Sememes are implicit in words and hence cannot be utilized in natural language processing (NLP) directly.
To tackle this challenge, \citet{dong2006hownet} manually defined about $2,000$ sememes and used them to annotate over $100,000$ English and Chinese words, whereupon a sememe knowledge base called HowNet was established.
Figure \ref{fig:hownet} gives an example of sememe annotations in HowNet.

HowNet is a special lexical knowledge base (KB).
Different from other lexical KBs like WordNet \citep{miller1998wordnet}, which explain meanings of words by relations between words, e.g., hyponym and meronym, HowNet provides intensional definitions of words using infra-word sememes.
This distinctness gives HowNet unique advantages.
First, sememes can be easily incorporated into neural networks as semantic labels \citep{qi2019modeling,qin2020improving}, which displays the particular suitability of HowNet in knowledge integration into deep learning. 
Second, the nature that limited sememes can represent meanings of unlimited words endows HowNet with the ability to handle low-data regimes, e.g., 
sememes can improve the embeddings of rare words 
 \citep{niu2017improved}. 
Because of these advantages, HowNet has been successfully utilized 
in various NLP tasks \citep{qi2021sememe}.

HowNet is distinctive and useful, but it covers only two languages (English and Chinese). Plus there are no sememe KBs like HowNet in other languages, which hinders NLP of most languages from benefiting by sememes. 
Manually building a sememe KB for each language is an obvious solution.  
But it is not realistic at all because the building process would be extremely time-consuming and labor-intensive --- it takes several linguistic experts more than two decades to build HowNet. 

\begin{figure}[t]
\centering
	\includegraphics[width=\linewidth]{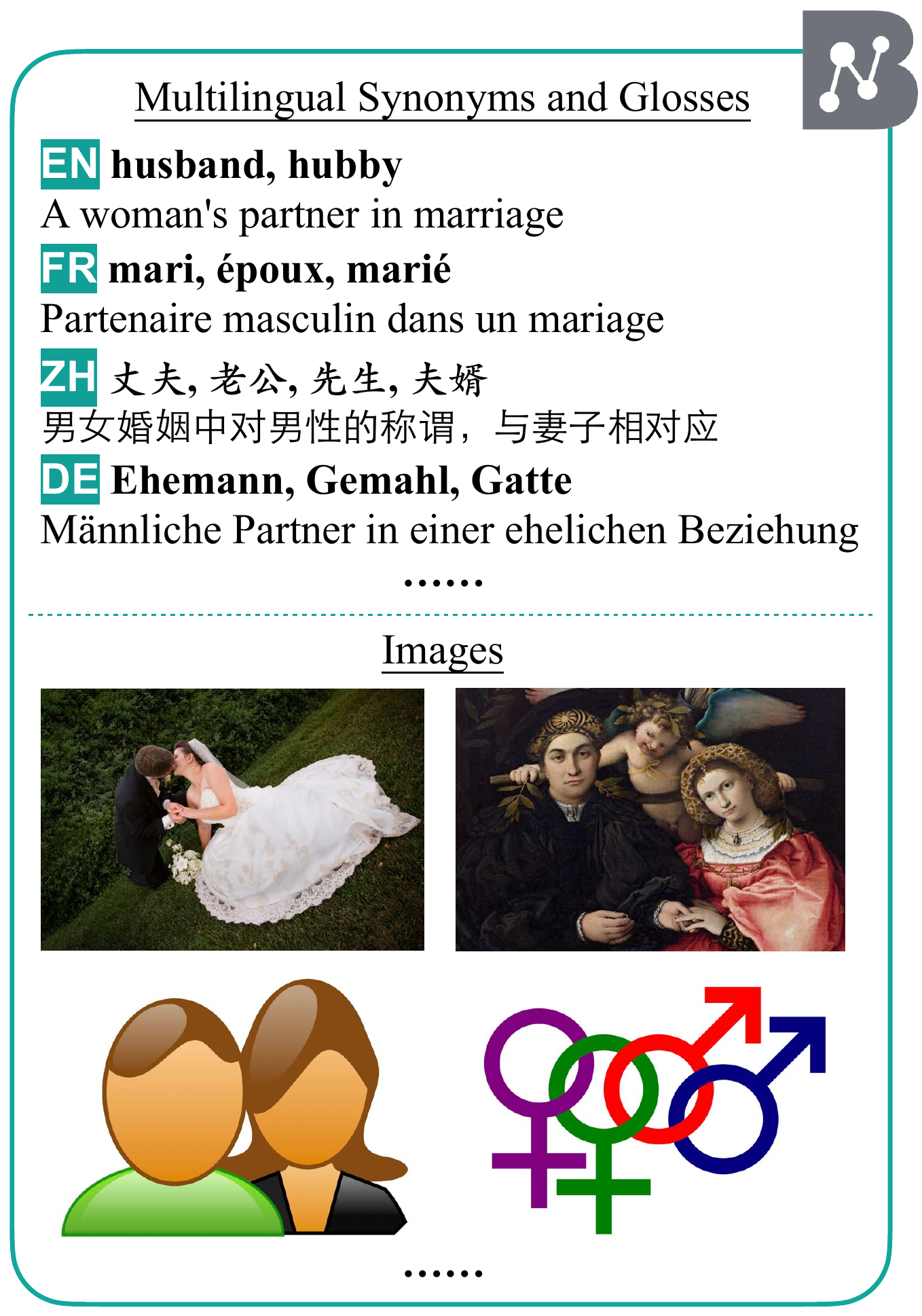}
	\caption{A BabelNet synset that comprises multilingual synonyms and glosses as well as some images.}
	\label{fig:babelnet}
\end{figure}

To solve this problem, \citet{qi2020towards} pioneeringly propose to build a \textit{multilingual} sememe KB based on BabelNet \citep{navigli2012babelnet}, a multilingual encyclopedic dictionary.
The entries of BabelNet are synsets composed of synonyms in almost 500 languages, as illustrated in Figure \ref{fig:babelnet}.
All the multilingual synonyms in a synset have the same meaning and thus should be annotated with the same sememes.
Therefore, sememe annotations of words in many languages would be simultaneously obtained by annotating sememes for BabelNet synsets. 
For example, suppose we annotate four sememes \texttt{human}, \texttt{family}, \texttt{spouse} and \texttt{male} to the synset in Figure \ref{fig:babelnet}, all the multilingual synonyms in the synset (``husband'', ``époux'', ``丈夫'', etc.) would be simultaneously annotated with these sememes.\footnote{If a word is polysemous, it would be included in multiple BabelNet synsets and have multiple sets of sememes.} 


Further, \citet{qi2020towards} build a seed dataset by manually annotating sememes for some synsets, and present the task of sememe prediction for BabelNet synsets (SPBS), which is aimed at automatically predicting sememes for the other unannotated synsets. 
In addition, they put forward two SPBS methods that utilize different information in BabelNet synsets, namely synset-related Wikipedia articles and relations between synsets.  


In this paper, we argue that some other information contained in BabelNet can be exploited for SPBS. 
As shown in Figure \ref{fig:babelnet}, in addition to the multilingual synonyms, each BabelNet synset comprises multilingual glosses that are extracted from different sources including WordNet and Wiktionary.\footnote{\url{https://www.wiktionary.org/}}
Besides, many synsets contain images from Wikipedia and Wikidata \citep{vrandevcic2014wikidata}.
The multilingual synonyms, glosses and images of a synset convey the meaning of the synset, thus naturally helpful in predicting sememes for the synset.
Therefore, we propose to utilize all the information in BabelNet synsets for the task of SPBS.

We design a multimodal information fusion model
named MSGI (sememe prediction with Multilingual Synonyms and Glosses as well as Images), which comprises a multilingual text encoder, an image encoder and a multi-label classifier. 
The text encoder is based on a cross-lingual pre-trained language model that encodes the multilingual synonyms and glosses. 
To adapt the general pre-trained language model for the task of SPBS, we introduce a new pre-training task named masked contextual sememe prediction to it.
The image encoder learns the embeddings of the images, and we adopt the attention-based multi-instance learning mechanism to process multiple images.

In experiments, we find that our MSGI model substantially outperforms previous SPBS methods (by about 10 MAP and F1 scores).
We also conduct a series of quantitative and qualitative analyses of the sememe prediction results of MSGI, aiming to explain the effectiveness of MSGI.

\section{Related Work}

\subsection{Sememe Knowledge Base}
HowNet is the most famous sememe KB and has attracted wide attention since it was published \citep{dong2006hownet}.
So far it has displayed its effectiveness in various NLP tasks, such as word similarity computation \citep{liu2002word}, sentiment analysis \citep{fu2013multi}, word sense disambiguation \citep{hou2020try}, word representation learning \citep{niu2017improved}, language modeling \citep{gu2018language}, relation extraction \citep{li2019chinese}, reverse dictionary \citep{zhang2020multi}, textual adversarial and backdoor attacks \citep{zang2020word,qi2021turn}, text matching \citep{lyu2021let}, quote recommendation \citep{qi2022quoter}, etc.

Besides the application of sememe KBs, another line of research is the automatic expansion and construction of sememe KBs. 
Among these studies, most of them focus on automatic expansion of existing sememe KBs \citep{xie2017lexical,jin2018incorporating,lyu2021glyph}. 
They propose different methods to automatically predict sememes for English/Chinese words that are not covered in HowNet, aiming to expand and update HowNet.

Only a few studies try to automatically construct a sememe KB for a new language.
\citet{qi2018cross} present the task of cross-lingual lexical sememe prediction, aiming to predict sememes for words in a new language based on the sememe annotations of English/Chinese words in HowNet.
However, it is not efficient because it can handle only one language at a time. 
Moreover, it cannot conduct sense-level sememe prediction and thus hardly processes polysemous words.
Afterwards, \citet{qi2020towards} pioneeringly propose the scheme of the BabelNet-based multilingual sememe KB, which is a more efficient and economical way to build sememe KBs for many languages. 
They take advantage of the multilingual nature of BabelNet and try to automatically predict sememes for all BabelNet synsets, so that all words in almost 500 languages in BabelNet would obtain sememe annotations.
Further, they build a seed dataset by aligning the words in HowNet and BabelNet and propose two methods to automatically predict sememes for synsets. 
Building on this work, we utilize more information incorporated in BabelNet to predict sememes for BabelNet synsets, achieving much better results.

Moreover, a recent work tries to construct a sememe KB on the basis of a dictionary \citep{qi2021automatic}.
It does not rely on the existing sememe annotations of HowNet or use the sememe set of HowNet.
Instead, it views the words in the controlled defining vocabulary of a dictionary as ``sememes'', and extracts them directly from dictionary definitions.

\subsection{BabelNet}
BabelNet \citep{navigli2012babelnet} is a multilingual encyclopedic dictionary that merges many heterogeneous resources, mainly including WordNet \citep{miller1998wordnet}, Wikipedia and Wikidata \citep{vrandevcic2014wikidata}.
It has been utilized in multiple NLP tasks \citep{navigli2021ten}, especially the cross-lingual or multilingual tasks, such as multilingual word sense disambiguation \citep{navigli2012joining}, cross-lingual lexical entailment \citep{vyas2016sparse} and cross-lingual AMR parsing \citep{blloshmi2020enabling}.
Most of these studies regard BabelNet as a large multilingual sense inventory and utilize the multilingual synonyms and glosses in BabelNet synsets, and some studies also use images in it, e.g., \citet{calabrese2020evilbert}  learn multimodal sense embeddings with the concepts and images in BabelNet.

Due to the multilingual mapping between different resources, BabelNet has become the hub to ground many linguistic resources, e.g., BabelNet is at the core of a dictionary matrix  within the ELEXIS project\footnote{\url{https://elex.is/}} that aims to interlink different lexicographic resources.


\section{Methodology}
\begin{figure*}[t]
\centering
	\includegraphics[width=\linewidth]{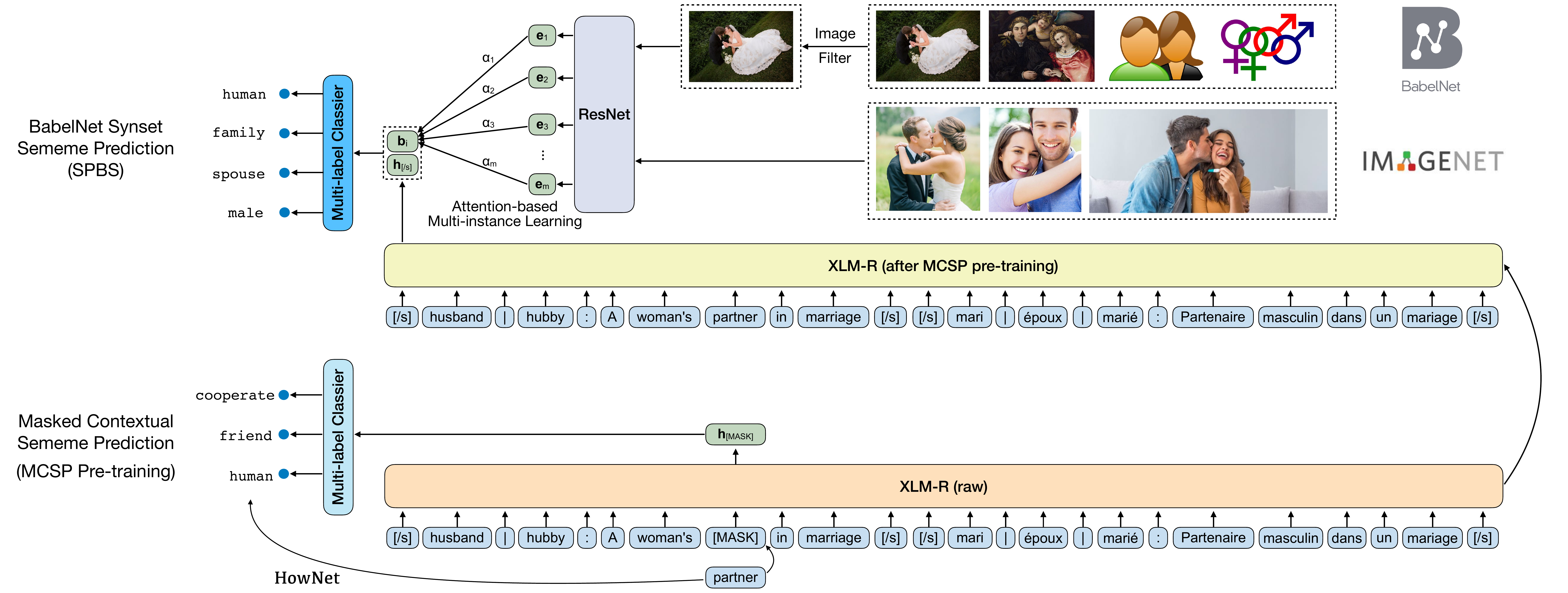}
	\caption{The illustration of the MSGI model. For simplicity, we only show the synonyms and glosses in two languages (English and French) in the multilingual text sequence.}
	\label{fig:method}
\end{figure*}

In this section, we elaborate on our MSGI model.
Before that, we first introduce the formalization of the SPBS task.
Then we describe the details of MSGI, and finally we present the training strategy of MSGI.
Figure \ref{fig:method} illustrates the framework and training strategy of MSGI.

\subsection{SPBS Task Formalization}
According to \citet{qi2020towards}, SPBS neglects the hierarchical structures of sememes and regards sememes as discrete semantic labels.
Therefore, SPBS is essentially a multi-label classification problem that is aimed at attaching appropriate labels (sememes) to the target BabelNet synset. 
Formally, suppose $\mathbb{B}$ is the set of all BabelNet synsets and $\mathbb{S}$ is the set of all sememes. 
For a given target synset $b\in \mathbb{B}$, SPBS is intended to predict its sememe set $\mathbb{S}_b=\{s_1, \cdots, s_{|\mathbb{S}_b|}\}\subset \mathbb{S}$, where $|\cdot|$ represents the cardinality of a set.

To this end, a prediction score is computed for each sememe.
Then the sememes whose prediction scores are higher than a threshold are selected as the prediction results. 
Formally, the predicted sememe set for the target synset $b$ is
\begin{equation}
\hat{\mathbb{S}}_b=\{s\in\mathbb{S}|P(s|b) > \delta \},
\label{eq:formalization}
\end{equation}
where $P(s|b)$ is the prediction score of a sememe $s$ and $\delta$ is the prediction score threshold.


\subsection{The MSGI Model}
MSGI is a multimodal information fusion model that is composed of a text encoder, an image encoder and a multi-label classifier. 
Next, we describe the three parts one by one.

\subsubsection*{Text Encoder}

The text encoder is aimed at encoding the semantic information of the multilingual synonyms and glosses of a BabelNet synset.
We combine all the multilingual synonyms and glosses into a multilingual text sequence and utilize XLM-R \citep{conneau2020unsupervised} to encode it. 
XLM-R is a large cross-lingual pre-trained language model, and is pre-trained on a large corpus in many languages using self-supervised training objectives including masked language model \citep{devlin2019bert}. 
Because of the popularity and outstanding performance on multiple cross-lingual NLP tasks, we choose XLM-R as the base text encoder in this paper. 
But our method also works based on other cross-lingual pre-trained language models.

We construct the multilingual text sequence of a synset in the following way.
For a target synset, we first concatenate the synonyms and gloss in the same language. 
Inspired by \citet{du2020lexical}, we put a special separator token, specifically a colon (:), between the synonyms and gloss to discriminate them.
Besides, we use another separator token, namely vertical bar ($|$), to separate the synonyms.
For example, the concatenation of the English synonyms and gloss of the example synset in Figure \ref{fig:babelnet} is \{\texttt{[/s]} \textit{husband $|$ hubby : A woman's partner in marriage} \texttt{[/s]}\}, where \texttt{[/s]} is the language separator token in XLM-R.

After obtaining the monolingual text sequences in many languages, we concatenate them into the final multilingual text sequence.
For example, the concatenation of the English and French text sequences is $S_{\{\text{en,fr}\}}$=\{\texttt{[/s]} \textit{husband $|$ hubby : A woman's partner in marriage} \texttt{[/s]} \texttt{[/s]} \textit{mari $|$ époux $|$ marié : Partenaire masculin dans un mariage} \texttt{[/s]}\}, as shown in Figure \ref{fig:method}.

Next, we feed the multilingual text sequence into XLM-R and obtain a series of hidden states:
\begin{equation}
	\mathbf{h}_{\texttt{[/s]}},\cdots=\text{XLM-R}({S}).
\end{equation}
We use the first hidden state as the text-based semantic representation of the synset: $\mathbf{b}_t=\mathbf{h}_{\texttt{[/s]}}$.

\subsubsection*{Image Encoder}
The image encoder is used to capture the semantic information contained in the images in a BabelNet synset.
Previous studies have shown that images can help learn better semantic representations for concepts and entities \citep{xie2017image,calabrese2020evilbert}.
We believe that images are also beneficial to SPBS.

We use the popular image classification model ResNet \citep{deng2009imagenet} as the image encoder to learn image embeddings. 
Most BabelNet synsets have multiple images, and we need to combine the embeddings of multiple images into one aggregated image-based representation.
Simply averaging all image embeddings may suffer from noises and cannot highlight important information.
Inspired by \citet{xie2017image}, we utilize the attention-based multi-instance learning mechanism to construct the aggregated image-based representation.

Suppose a BabelNet synset $b$ has $m$ images and the embedding of the $j$-th image obtained from RestNet is $\mathbf{e}_j$. 
Based on the text-based representation of the synset $\mathbf{b}_t$, we calculate the attention of each image:
\begin{equation}
	\alpha_j=\frac{\exp(\mathbf{b}_t\cdot\mathbf{e}_j)}{\sum_{k=1}^m \exp(\mathbf{b}_t\cdot\mathbf{e}_k)}.
\end{equation}
The aggregated image-based representation is the attention-weighted sum of the image embeddings: $\mathbf{b}_i=\sum_{j=1}^m \alpha_j\mathbf{e}_j$.

In experiments, however, we find that images in BabelNet are too diversified, and some are even not related to the corresponding synsets at all.\footnote{It is because most images in BabelNet are automatically extracted from Wikipedia and Wikidata without manual examination.}
For example, among the displayed four images in the example synset in Figure \ref{fig:babelnet}, they vary markedly in styles and semantic descriptive perspectives. 
Even with the attention mechanism, the model would still be confused if we consider all the images.

To tackle this issue, we take the following two measures: 
(1) \textbf{Removing Low-quality Images}. We adopt an unsupervised outlier detection algorithm, more specifically One-Class-SVM \citep{scholkopf1999support}, to detect and filter out some low-quality images based on their image embeddings;
(2) \textbf{Adding High-quality Images}. Since BabelNet synsets are connected with WordNet synsets, we can retrieve more images for some BabelNet synsets from ImageNet \citep{deng2009imagenet} that is also organized based on WordNet. Images in ImageNet are manually annotated and have much higher quality.
After the two measures, we obtain a better image set, and then we adopt the attention-based multi-instance learning mechanism to obtain the final image-based representation $\mathbf{b}_i$.

\subsubsection*{Multi-label Classifier}

We concatenate the text-based and image-based representations of a synset, and pass the concatenation vector into a single-layer perceptron for multi-label classification: 
\begin{equation}
	\mathbf{p}=\sigma(\mathbf{W}[\mathbf{b}_t;\mathbf{b}_i]+\boldsymbol{\mu}),
	\label{eq:cls}
\end{equation}
where $\mathbf{W}$ is a weight matrix, $\boldsymbol{\mu}$ is a bias vector, and $\sigma$ is the sigmoid function.
The obtained $\mathbf{p}\in\mathbb{R}^{|\mathbb{S}|}$ is the sememe prediction score vector whose $i$-th element is the prediction score of the $i$-th sememe.

\subsection{Training Strategy of MSGI} 

We can simply train MSGI using the cross-entropy loss, during which the text encoder (XLM-R) is fine-tuned, the multi-label classifier is trained, but the image encoder (ResNet) is frozen.\footnote{We find that freezing rather than tuning ResNet can obtain higher performance, presumably because of the limited size of the training set, which is consistent with the findings in previous studies \citep{xie2017image}.}
The training loss of a training instance $b$ is
\begin{equation}
	\mathcal{L}_b=-\frac{1}{\mathbb{S}}\big[\sum_{s\in \mathbb{S}_b}\log p_s+\sum_{s\notin \mathbb{S}_b}\log(1-p_s)\big],
	\label{eq:loss}
\end{equation}
where $p_s$ is the sememe prediction score of $s$.

Here we directly use the raw XLM-R, which is general and independent on downstream tasks.
We argue that it can be enhanced by integrating specific adaptation to the SPBS task.
Inspired by the masked language model \citep{devlin2019bert} and sememe-incorporated language model \citep{gu2018language}, we propose the masked contextual sememe prediction (MCSP) pre-training task as the adaptation of XLM-R.

\subsubsection*{MCSP Pre-training}
MCSP is aimed at predicting sememes for a masked word in a sentence by utilizing the contextual information.
It is viable for English and Chinese glosses thanks to HowNet that annotates sememes for English and Chinese words.
We hope that MCSP pre-training can make the raw XLM-R more familiar with sememes and in turn, perform better in the subsequent training of SPBS.

More specifically, 
for a multilingual text sequence of a synset, we randomly replace some words in its English and Chinese glosses with a special \texttt{[MASK]} token.
Then we feed the corrupted text sequence into the raw XLM-R, and pass the hidden states of the \texttt{[MASK]} tokens to a multi-label classifier like Equation \eqref{eq:cls}, which serves as the sememe predictor for words.
Following previous studies on sememe prediction for words \citep{xie2017image,jin2018incorporating}, we neglect the polysemy of the masked words and group sememes of all senses together to form the sememe set of a word.

The training loss for MCSP is also multi-label cross-entropy loss.
After the MCSP pre-training, we conduct the training of SPBS as in Equation \eqref{eq:loss}.



\begin{table*}[!t]
\centering
\resizebox{.89\linewidth}{!}{%
\begin{tabular}{c|cccccccc|cc}
\toprule
\multicolumn{1}{c|}{PoS \small{(\#synset)}} & \multicolumn{2}{c}{Noun \small{(10,360)}} & \multicolumn{2}{c}{Verb \small{(2,240)}} & \multicolumn{2}{c}{Adj. \small{(2,419)}} & \multicolumn{2}{c}{Adv. \small{(442)}} & \multicolumn{2}{|c}{All \small{(15,461)}} \\ 
\midrule
\multicolumn{1}{c|}{Model} & MAP & F1 & MAP & F1 & MAP & F1 & MAP & F1 & MAP & F1 \\ 
\midrule
LR & 54.42 & 39.81 & -- & -- & -- & -- & -- & -- & -- & -- \\
TransE & 61.05 & 46.78 & 34.75 & 26.76 & 29.11 & 22.99 & 30.05 & 20.69 & 51.73 & 39.73 \\ 
\midrule
SPBS-SR & 65.16 & 49.75 & -- & -- & -- & -- & -- & -- & -- & -- \\
SPBS-RR & 62.50 & 47.92 & 34.76 & 25.28 & 32.68 & 24.51 & 30.86 & 20.07 & 53.31 & 40.53 \\
Ensemble & 68.85 & 55.35 & 34.76 & 25.28 & 32.68 & 24.51 & 30.86 & 20.07 & 57.64 & 45.61 \\
\midrule
MSGI (ours) & \textbf{71.81} & \textbf{64.36} & \textbf{59.78} & \textbf{47.01} & {55.61} & {41.02} & {68.52} & \underline{55.20} & \textbf{67.23} & \textbf{57.68} \\
\multicolumn{1}{r|}{-Synonym} & 67.40 & 59.07 & 35.31 & 24.99 & 36.33 & 26.18 & 48.33 & 37.45 & 57.25 & 48.54 \\ 
\multicolumn{1}{r|}{-Gloss} & 66.90 & 56.99 & 54.22 & 41.54 & 53.11 & 39.20 & \textbf{68.76} & \underline{55.14} & 62.67 & 52.21 \\ 
\multicolumn{1}{r|}{-Image} & 71.41 & 61.58 & 59.70 & 44.29 & \textbf{55.86} & \textbf{43.15} & 63.81 & 51.63 & 67.13 & 56.62 \\ 
\multicolumn{1}{r|}{-MCSP} & 70.58 & 61.99 & 57.55 & 43.27 & 52.57 & 40.61 & 68.49 & 52.79 & 65.70 & 56.05 \\
\bottomrule
\end{tabular}%
}
\caption{SPBS performance of different models on the test set of BabelSememe. The boldfaced \textbf{results} exhibit statistically significant improvement over the other results with $p<0.1$ according to the paired $t$-test, and the underlined \underline{results} indicate no significant difference.}
\label{tab:main}

\end{table*}

\section{Experiments}
In this section, we evaluate the sememe prediction performance of our MSGI model.

\subsection{Experimental Settings}
\paragraph{Dataset}
The evaluation is conducted on BabelSememe, the seed dataset of the multilingual sememe KB based on BabelNet that is built by \citet{qi2020towards}.
Its training/validation/test sets have $12,369$/$1,546$/$1,546$ synsets that are manually annotated by a total of $2,106$ sememes.

\paragraph{Baseline Methods}
We choose the two methods proposed by \citet{qi2020towards} as main baselines: 
(1) \textbf{SPBS-SR}, which 
performs collaborative filtering-based sememe prediction \citep{xie2017lexical} using NASARI embeddings \citep{camacho2016nasari}, a set of synset embeddings trained with related Wikipedia articles;
(2) \textbf{SPBS-RR}, which models SPBS as a relation prediction task in knowledge graph by considering relations between synsets;
(3) the \textbf{Ensemble} of the above two methods.
Besides, we have two naive baselines that are used for comparison in \citet{qi2020towards};
(4) Logistic regression (\textbf{LR}), which directly uses  NASARI embeddings for multi-label classification;
(5) \textbf{TransE} \citep{bordes2013translating}, which is a classical relation prediction model and adapted for SPBS in a similar way to SPBS-RR.\footnote{SPBS-SR and LR require NASARI embeddings that only cover nominal synsets. Thus the two methods work on the nominal synsets only.}

\paragraph{Evaluation Metrics}
Following \citet{qi2020towards}, we use mean average precision (MAP) and F1 score as the evaluation metrics.

\paragraph{Selection of Languages}
It is impractical to consider all the 500 languages in BabelNet together. 
In our experiments, we pick 3 representative languages, namely English, French and Chinese.
English and Chinese are the two languages in HowNet and are required for MCSP pre-training.
French is a high-resource language and most synsets have French glosses in BabelNet.
Besides, these 3 languages have different linguistic distances: English is close to French while Chinese is far from the two.
Some synsets have no glosses in French or Chinese, and we remove the whole corresponding monolingual part from the multilingual text sequences.

\paragraph{Implementation Details}
For the text encoder, we use the pre-trained base version of XLM-R with the help of the Transformers library \citep{wolf2020transformers}, and the hidden size is 768.
For the image encoder, we choose ResNet-152 that contains 152 layers and delivers 1000-dimensional image embeddings, and implement the model with PyTorch.\footnote{\url{https://pytorch.org/hub/pytorch_vision_resnet/}}
We transform the image embeddings into 768 dimensions with a linear layer in order for attention calculate and concatenation with the text-based representation.
For images from BabelNet, we resize them into 256$\times$256.
For images from ImageNet, we use the processed version of ImageNet 21K \citep{ridnik2021imagenet} whose images are resized into 224$\times$224.
In BabelSememe, 9,356 synsets have images, among which 2,538 synsets have images from both BabelNet and ImageNet. 
The average image number of a synset is 45.

We use the Adam \citep{kingma2014adam} optimizer in both MCSP pre-training and the final training. 
The prediction score threshold $\delta$ in Equation \eqref{eq:formalization} is continuously tuned on the validation set and set to 0.42 finally.
The learning rates for XLM-R and the multi-label classifier are separately tuned in \{1e-6, 5e-6, \textbf{1e-5}, 5e-5, 1e-4\} and \{1e-4, 5e-4, \textbf{1e-3}, 5e-3, 1e-2\}, where the boldfaced ones are final picks based on the validation set performance.

\subsection{Main Results}
\label{sec:main_results}

Table \ref{tab:main} shows the SPBS results of different models on the test set. 
We have the following observations:

(1) The MSGI model achieves consistent and substantial outperformance over previous methods (about 10 for both MAP and F1 score), which demonstrates the usefulness of the multilingual and multimodal information in BabelNet in the SPBS task and the effectiveness of the MSGI model.

(2) Among the four PoS types, MSGI performs best on nominal synsets, which is possibly because nominal synsets have the largest amount and the most abundant information in BabelNet \citep{navigli2012babelnet}.

(3) MSGI largely improves the performance on the non-nominal synsets compared with TransE and SPBS-RR. 
It is because the baselines rely on the relations between synsets, and non-nominal synsets have sparse relations \citep{qi2020towards}.
In contrast, MSGI utilizes the internal information of BabelNet synsets and is immune to the relation density.

\subsubsection*{Ablation Study}
We conduct a series of ablation studies to show the effectiveness of different parts of the MSGI model.
(1) \textbf{-Synonym}. We eliminate all the synonyms and separator tokens in the multilingual text sequences, i.e., retain the glosses only.
(2) \textbf{-Gloss}. We remove all the multilingual glosses and the colon separator tokens, and keep the synonyms together with the vertical bar separator tokens only.
(3) \textbf{-Image}. We remove the image encoder and use the text encoder together with the multi-label classifier only.
(4) \textbf{-MCSP}. We skip the MCSP pre-training and directly train the MSGI model on the raw XLM-R.

The results are also shown in Table \ref{tab:main}.
We can see that the original MSGI model has better overall results than all the above four incomplete models, which proves the effectiveness of the four parts.

\begin{table}
\centering
\resizebox{\linewidth}{!}{%
\begin{tabular}{lcc}
\toprule
Used Images & MAP & F1 \\ 
\midrule
No Images & 69.40 & 60.44 \\
All BabelNet Images & 70.25  & 60.99  \\
Filtered BabelNet Images & 70.63 & 61.21 \\
Filtered BabelNet + ImageNet Images & \textbf{71.33} & \textbf{62.10} \\

\bottomrule
\end{tabular}%
}
\caption{SPBS performance of the MSGI model incorporated with different image information.}
\label{tab:image}
\end{table}

\subsection{Effectiveness of Image Encoding}
According to the ablation study, the benefit of the images seems to be marginal. 
We conjecture that it is because many synsets (6,105, $\sim$40\%) have no available images and the image encoder only plays a limited role.
To better demonstrate the effectiveness of image encoding, we conduct experiments on the 9,356 synsets with images, which are randomly split into the training, validation and test sets in the ratio of 8:1:1.
In addition, we investigate the effectiveness of the two measures in image encoding, i.e., filtering BabelNet images and adding ImageNet images, on this subset.

Table \ref{tab:image} shows the results. 
We can see that the improvement brought by image encoding is better exhibited (nearly 2 MAP and F1 scores).
Further, both the two measures in image encoding are effective and improve the SPBS performance.

\subsection{Effectiveness of Multilinguality}
In this subsection, we investigate the effectiveness of the multilingual information in the MSGI model.
We extract the 8,974 synsets that have synonyms and glosses in all the three languages (English, French and Chinese), and randomly split them into training/validation/test sets in the ratio of 8:1:1.
Then we train MSGI with multilingual text sequences in different combinations of languages.

The evaluation results on the test set are shown in Table \ref{tab:language}.
We observe that considering more languages can bring performance enhancement indeed, which demonstrates the usefulness of the multilingual information in the SPBS task. 
We conjecture the possible reason is that the text sequences in different languages provide semantic information from different perspectives, and combining them can obtain more semantic information to better predict sememes. 
Besides, En+Zh and Fr+Zh outperform En+Fr, which indicates that the combination of distant languages can produce more benefits, presumably because text sequences in distant languages have more different semantic information.

\begin{table}
\centering
\resizebox{.58\linewidth}{!}{%
\begin{tabular}{lcc}
\toprule
Languages & MAP & F1 \\ 
\midrule
En & 67.22 & 55.80 \\
Fr & 59.87  & 50.87  \\
Zh & 70.87 & 61.13 \\
En+Fr & 68.01 & 57.48 \\
En+Zh & 71.95 & 61.53 \\
Fr+Zh & 71.65 & 60.45 \\
En+Fr+Zh & \textbf{72.98} & \textbf{63.46} \\ \bottomrule
\end{tabular}%
}
\caption{SPBS performance of the MSGI model with information in different language combinations.}
\label{tab:language}
\end{table}

\begin{figure}[t]
\centering
	\includegraphics[width=.85\linewidth]{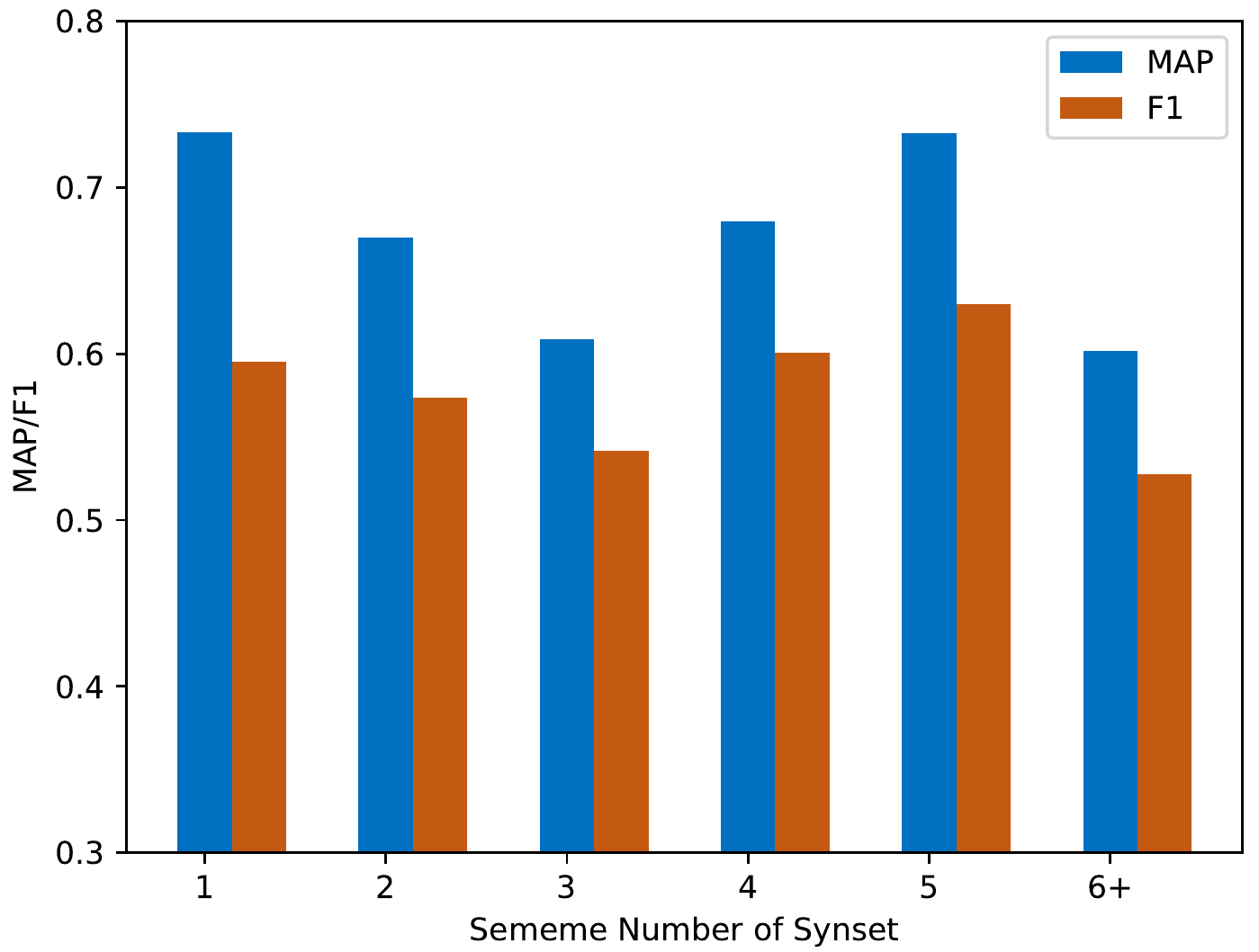}
	\caption{SPBS results of synsets with different numbers of sememes. The numbers of synsets in the six ranges are 422, 422, 287, 208, 119 and 88, respectively.}
	\label{fig:sememe-num}
\end{figure}

\section{Analysis}
In this section, we conduct some quantitative and qualitative analyses of the SPBS results of MSGI.
All the experiments are conducted on the validation set of BabelSememe.

\subsection{Effect of Synset's Sememe Number}
We first investigate how the characteristics of a synset affect its sememe prediction results.
The effect of PoS has been studied in §\ref{sec:main_results}. 
Here we focus on another quantitative characteristic, namely the number of a synset's annotated sememes.
Figure \ref{fig:sememe-num} shows the average sememe prediction MAP and F1 scores of the synsets that have different numbers of sememes. 
We find that the sememe prediction performance of a synset is basically not influenced by its sememe number.
In contrast, according to \citet{qi2020towards}, the baseline methods (SPBS-SR, SPBS-RR and Ensemble) perform badly on the synsets with too few or too many sememes.
These results show the higher robustness of our MSGI model to sememe number.

\begin{figure}[t]
\centering
	\includegraphics[width=.85\linewidth]{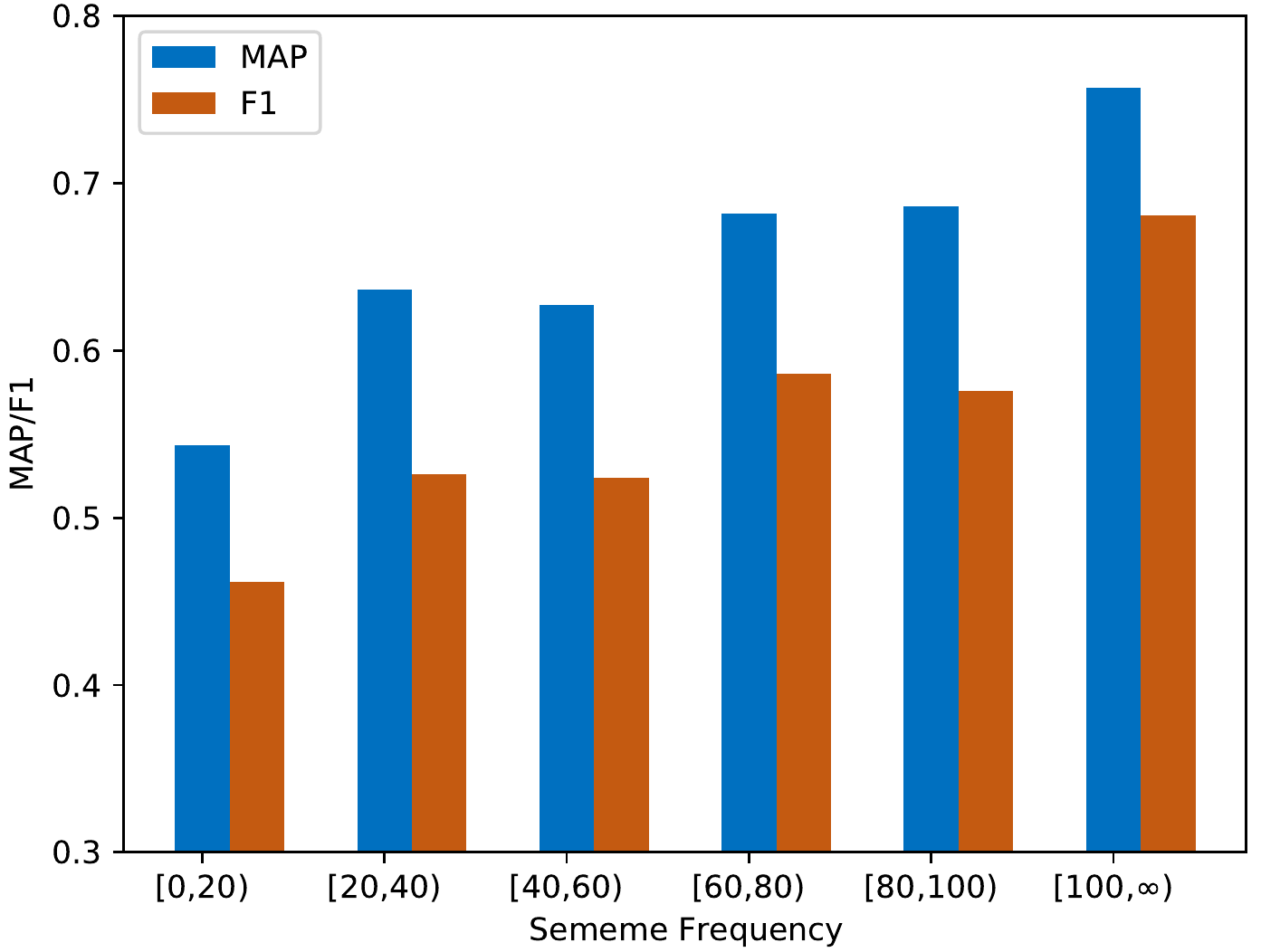}
	\caption{SPBS results of synsets having sememes with different frequencies. The numbers of synsets in the six ranges are 708, 164, 66, 35, 21 and 49, respectively.}
	\label{fig:sememe-freq}
\end{figure}

\subsection{Effect of Sememe Frequency}
In this subsection, we explore what sememes are easy or hard to predict.
We study the characteristic of sememe frequency, i.e., the number of synsets having a target sememe in the training set, which is the only quantitative feature of sememes.
Figure \ref{fig:sememe-freq} shows the results, where the x-axis denotes the sememe frequency ranges while the y-axis denotes the average sememe prediction performance of the synsets having the sememes within a specific frequency range.
We find that the frequent sememes are easier to predict broadly, which is consistent with the findings in previous work \citep{qi2020towards}.

\begin{table*}[t!]
\centering
\resizebox{\textwidth}{!}{%
\begin{tabular}{@{}c|ll|l|l@{}}
\toprule
\multirow{2}{*}{Type} & \multicolumn{2}{c|}{Example Synset} & \multicolumn{1}{c|}{\multirow{2}{*}{Predicted Sememes}} & \multicolumn{1}{c}{\multirow{2}{*}{Ground Truth}} \\ 
\cline{2-3}
& \multicolumn{1}{c|}{Synonym} & \multicolumn{1}{c|}{Gloss} & &  \\ 
\midrule
Good & \multicolumn{1}{c|}{Egypt} & \makecell[l]{A republic in northeastern Africa} & \makecell[l]{\textbf{politics}, \textbf{place}, \textbf{country}, \\ \textbf{ProperName}, \textbf{Africa}} & \makecell[l]{politics, place, country, \\ ProperName, Africa} \\
\hline
Fewer &  \multicolumn{1}{c|}{anorexia} & \makecell[l]{A psychological disorder characterized by somatic delusions \\ that you are too fat despite being emaciated} & \makecell[l]{\textbf{disease}} & \makecell[l]{disease, disgust, eat} \\
\hline
More & \multicolumn{1}{c|}{boiler} & \makecell[l]{A pressurized system in which water is vaporized to steam \\ by heat transferred from a source of higher temperature} & \makecell[l]{StateChange, produce, industrial, \\ \textbf{burn}, cook, \textbf{WarmUp}, \textbf{tool}} & \makecell[l]{burn, WarmUp, tool} \\ 
\hline
Similar &  \multicolumn{1}{c|}{semantic} & \makecell[l]{Of or relating to meaning or the study of meaning} & \makecell[l]{\textbf{language}, knowledge} & \makecell[l]{language, information} \\
\bottomrule
\end{tabular}%
}
\caption{Example synsets of four types classified according to sememe prediction results. We only show one English synonym and gloss for succinctness. The boldfaced \textbf{sememes} are the correctly predicted ones.}
\label{tab:case}
\end{table*}

\subsection{Qualitative Analysis}
In this subsection, we conduct qualitative analysis and case studies into the SPBS results of the MSGI model.
We randomly select fifty synsets from the validation set, and carry out error analysis one by one.
According to their sememe prediction results, we can classify the synsets into four types, namely (1) \textbf{Good}: MSGI performs well on these synsets with MAP/F1 score higher than 85; (2) \textbf{Fewer}, MSGI predicts fewer sememes for these synsets than the ground truth; (3) \textbf{More}, MSGI predicts more sememes for these synsets than the ground truth; (4) \textbf{Similar}: MSGI predicts some sememes that are different from but similar to the ground-truth sememes. 
The number of synsets belonging to the four types are 23 (46\%), 10 (20\%), 3 (6\%) and 14 (28\%), respectively.

We pick one example synset for each type and show their basic information and sememe prediction results in Table \ref{tab:case}.
For the synset of ``anorexia'', the gloss doesn't embody any information about ``disgust at eating'', thus the MSGI model doesn't predict the two sememes ``disgust'' and ``eat''.
For the synset of ``boiler'', the gloss provides much information and the model predicts more sememes than the ground truth, which are basically reasonable.
For the synset of ``semantic'', our model predicts ``knowledge'' rather than ``information'', while the two sememes are similar and related.

\section{Conclusion and Future Work}
In this paper, we propose to utilize the multilingual and multimodal information in BabelNet, i.e., multilingual synonyms, multilingual glosses and images, to predict sememes for BabelNet synsets. 
We design the MSGI model and it achieves absolute outperformance over previous methods.
In the future, we will try to leverage more information in BabelNet, e.g., semantic relations, to better predict sememes. 
We will also consider expanding BabelSememe with the prediction results of our model after manual examination. 

\section*{Acknowledgements}
This work is supported by the National Key R\&D Program of China (No. 2020AAA0106502), Institute Guo Qiang at Tsinghua University, and International Innovation Center of Tsinghua University, Shanghai, China. Zheng and Lv are supported by  National Natural Science Foundation of China (Grant No. 6201101015), Beijing Academy of Artificial Intelligence (BAAI)， Natural Science Foundation of Guangdong Province (Grant No. 2021A1515012640), the Basic Research Fund of Shenzhen City (Grant No. JCYJ20210324120012033 and JCYJ20190813165003837), and Overseas Cooperation Research Fund of Tsinghua Shenzhen International Graduate School  (Grant No. HW2021008).
We also thank all the anonymous reviewers for their valuable comments and suggestions.

\section*{Ethical Statements}
In this paper, we use only one dataset, namely BabelSememe, which is completely free and publicly available.
The task we tackle, namely SPBS, is only related to NLP research and not for practical application, thus cannot be misused by the ordinary people.
To save energy, we use the base version of XLM-R rather than larger cross-lingual pre-trained models, although they may yield higher performance.
No demographic or identity characteristics are used in the formation of this paper.

\bibliographystyle{acl_natbib}
\bibliography{custom}

\begin{thebibliography}{43}
\expandafter\ifx\csname natexlab\endcsname\relax\def\natexlab#1{#1}\fi

\bibitem[{Blloshmi et~al.(2020)Blloshmi, Tripodi, and
  Navigli}]{blloshmi2020enabling}
Rexhina Blloshmi, Rocco Tripodi, and Roberto Navigli. 2020.
\newblock Enabling cross-lingual amr parsing with transfer learning techniques.
\newblock In \emph{Proceedings EMNLP}, pages 2487--2500.

\bibitem[{Bloomfield(1926)}]{bloomfield1926set}
Leonard Bloomfield. 1926.
\newblock A set of postulates for the science of language.
\newblock \emph{Language}, 2(3):153--164.

\bibitem[{Bordes et~al.(2013)Bordes, Usunier, Garcia-Duran, Weston, and
  Yakhnenko}]{bordes2013translating}
Antoine Bordes, Nicolas Usunier, Alberto Garcia-Duran, Jason Weston, and Oksana
  Yakhnenko. 2013.
\newblock Translating embeddings for modeling multi-relational data.
\newblock \emph{Proceedings of NIPS}.

\bibitem[{Calabrese et~al.(2020)Calabrese, Bevilacqua, and
  Navigli}]{calabrese2020evilbert}
Agostina Calabrese, Michele Bevilacqua, and Roberto Navigli. 2020.
\newblock Evilbert: Learning task-agnostic multimodal sense embeddings.
\newblock In \emph{Proceedings of IJCAI}.

\bibitem[{Camacho-Collados et~al.(2016)Camacho-Collados, Pilehvar, and
  Navigli}]{camacho2016nasari}
Jos{\'e} Camacho-Collados, Mohammad~Taher Pilehvar, and Roberto Navigli. 2016.
\newblock Nasari: Integrating explicit knowledge and corpus statistics for a
  multilingual representation of concepts and entities.
\newblock \emph{Artificial Intelligence}, 240:36--64.

\bibitem[{Conneau et~al.(2020)Conneau, Khandelwal, Goyal, Chaudhary, Wenzek,
  Guzm{\'a}n, Grave, Ott, Zettlemoyer, and Stoyanov}]{conneau2020unsupervised}
Alexis Conneau, Kartikay Khandelwal, Naman Goyal, Vishrav Chaudhary, Guillaume
  Wenzek, Francisco Guzm{\'a}n, {\'E}douard Grave, Myle Ott, Luke Zettlemoyer,
  and Veselin Stoyanov. 2020.
\newblock Unsupervised cross-lingual representation learning at scale.
\newblock In \emph{Proceedings of ACL}.

\bibitem[{Deng et~al.(2009)Deng, Dong, Socher, Li, Li, and
  Fei-Fei}]{deng2009imagenet}
Jia Deng, Wei Dong, Richard Socher, Li-Jia Li, Kai Li, and Li~Fei-Fei. 2009.
\newblock Imagenet: A large-scale hierarchical image database.
\newblock In \emph{Proceedings of CVPR}.

\bibitem[{Devlin et~al.(2019)Devlin, Chang, Lee, and
  Toutanova}]{devlin2019bert}
Jacob Devlin, Ming-Wei Chang, Kenton Lee, and Kristina Toutanova. 2019.
\newblock {BERT}: Pre-training of deep bidirectional transformers for language
  understanding.
\newblock In \emph{Proceedings of NAACL-HLT}.

\bibitem[{Dong and Dong(2006)}]{dong2006hownet}
Zhendong Dong and Qiang Dong. 2006.
\newblock \emph{HowNet and the Computation of Meaning (With CD-Rom)}.
\newblock World Scientific.

\bibitem[{Du et~al.(2020)Du, Qi, Sun, and Liu}]{du2020lexical}
Jiaju Du, Fanchao Qi, Maosong Sun, and Zhiyuan Liu. 2020.
\newblock Lexical sememe prediction by dictionary definitions and local
  semantic correspondence.
\newblock \emph{Journal of Chinese Information Processing}, 34(5):1--9.

\bibitem[{Fu et~al.(2013)Fu, Liu, Guo, and Wang}]{fu2013multi}
Xianghua Fu, Guo Liu, Yanyan Guo, and Zhiqiang Wang. 2013.
\newblock Multi-aspect sentiment analysis for {Chinese} online social reviews
  based on topic modeling and {HowNet} lexicon.
\newblock \emph{Knowledge-Based Systems}, 37:186--195.

\bibitem[{Gu et~al.(2018)Gu, Yan, Zhu, Liu, Xie, Sun, Lin, and
  Lin}]{gu2018language}
Yihong Gu, Jun Yan, Hao Zhu, Zhiyuan Liu, Ruobing Xie, Maosong Sun, Fen Lin,
  and Leyu Lin. 2018.
\newblock Language modeling with sparse product of sememe experts.
\newblock In \emph{Proceedings of EMNLP}.

\bibitem[{Hou et~al.(2020)Hou, Qi, Zang, Zhang, Liu, and Sun}]{hou2020try}
Bairu Hou, Fanchao Qi, Yuan Zang, Xurui Zhang, Zhiyuan Liu, and Maosong Sun.
  2020.
\newblock Try to substitute: An unsupervised chinese word sense disambiguation
  method based on hownet.
\newblock In \emph{Proceedings of COLING}.

\bibitem[{Jin et~al.(2018)Jin, Zhu, Liu, Xie, Sun, Lin, and
  Lin}]{jin2018incorporating}
Huiming Jin, Hao Zhu, Zhiyuan Liu, Ruobing Xie, Maosong Sun, Fen Lin, and Leyu
  Lin. 2018.
\newblock Incorporating chinese characters of words for lexical sememe
  prediction.
\newblock In \emph{Proceedings of ACL}.

\bibitem[{Kingma and Ba(2015)}]{kingma2014adam}
Diederik~P Kingma and Jimmy Ba. 2015.
\newblock Adam: A method for stochastic optimization.
\newblock In \emph{Proceedings of ICLR}.

\bibitem[{Li et~al.(2019)Li, Ding, Liu, Zheng, and Shen}]{li2019chinese}
Ziran Li, Ning Ding, Zhiyuan Liu, Haitao Zheng, and Ying Shen. 2019.
\newblock Chinese relation extraction with multi-grained information and
  external linguistic knowledge.
\newblock In \emph{Proceedings of ACL}.

\bibitem[{Liu and Li(2002)}]{liu2002word}
Qun Liu and Sujian Li. 2002.
\newblock Word similarity computing based on {HowNet}.
\newblock \emph{International Journal of Computational Linguistics \& {Chinese}
  Language Processing}, 7(2):59--76.

\bibitem[{Lyu et~al.(2021{\natexlab{a}})Lyu, Chen, and Yu}]{lyu2021glyph}
Boer Lyu, Lu~Chen, and Kai Yu. 2021{\natexlab{a}}.
\newblock Glyph enhanced {C}hinese character pre-training for lexical sememe
  prediction.
\newblock In \emph{Findings of the ACL: EMNLP 2021}.

\bibitem[{Lyu et~al.(2021{\natexlab{b}})Lyu, Chen, Zhu, and Yu}]{lyu2021let}
Boer Lyu, Lu~Chen, Su~Zhu, and Kai Yu. 2021{\natexlab{b}}.
\newblock Let: Linguistic knowledge enhanced graph transformer for chinese
  short text matching.
\newblock In \emph{Proceedings of AAAI}.

\bibitem[{Miller(1998)}]{miller1998wordnet}
George Miller. 1998.
\newblock \emph{{WordNet}: An electronic lexical database}.
\newblock MIT press.

\bibitem[{Navigli et~al.(2021)Navigli, Bevilacqua, Conia, Montagnini, and
  Cecconi}]{navigli2021ten}
Roberto Navigli, Michele Bevilacqua, Simone Conia, Dario Montagnini, and
  Francesco Cecconi. 2021.
\newblock Ten years of babelnet: A survey.
\newblock In \emph{Proceedings IJCAI}.

\bibitem[{Navigli and Ponzetto(2012{\natexlab{a}})}]{navigli2012babelnet}
Roberto Navigli and Simone~Paolo Ponzetto. 2012{\natexlab{a}}.
\newblock Babelnet: The automatic construction, evaluation and application of a
  wide-coverage multilingual semantic network.
\newblock \emph{Artificial Intelligence}, 193:217--250.

\bibitem[{Navigli and Ponzetto(2012{\natexlab{b}})}]{navigli2012joining}
Roberto Navigli and Simone~Paolo Ponzetto. 2012{\natexlab{b}}.
\newblock Joining forces pays off: Multilingual joint word sense
  disambiguation.
\newblock In \emph{Proceedings of EMNLP}.

\bibitem[{Niu et~al.(2017)Niu, Xie, Liu, and Sun}]{niu2017improved}
Yilin Niu, Ruobing Xie, Zhiyuan Liu, and Maosong Sun. 2017.
\newblock Improved word representation learning with sememes.
\newblock In \emph{Proceedings of ACL}.

\bibitem[{O'Grady et~al.(1997)O'Grady, Dobrovolsky, and
  Katamba}]{o1997contemporary}
William O'Grady, Michael Dobrovolsky, and Francis Katamba. 1997.
\newblock \emph{Contemporary Linguistics}.

\bibitem[{Qi et~al.(2020)Qi, Chang, Sun, Ouyang, and Liu}]{qi2020towards}
Fanchao Qi, Liang Chang, Maosong Sun, Sicong Ouyang, and Zhiyuan Liu. 2020.
\newblock Towards building a multilingual sememe knowledge base: Predicting
  sememes for babelnet synsets.
\newblock In \emph{Proceedings of AAAI}.

\bibitem[{Qi et~al.(2021{\natexlab{a}})Qi, Chen, Wang, Liu, Chen, and
  Sun}]{qi2021automatic}
Fanchao Qi, Yangyi Chen, Fengyu Wang, Zhiyuan Liu, Xiao Chen, and Maosong Sun.
  2021{\natexlab{a}}.
\newblock Automatic construction of sememe knowledge bases via dictionaries.
\newblock In \emph{Findings of the Association for Computational Linguistics:
  ACL-IJCNLP 2021}.

\bibitem[{Qi et~al.(2019)Qi, Huang, Yang, Liu, Chen, Liu, and
  Sun}]{qi2019modeling}
Fanchao Qi, Junjie Huang, Chenghao Yang, Zhiyuan Liu, Xiao Chen, Qun Liu, and
  Maosong Sun. 2019.
\newblock Modeling semantic compositionality with sememe knowledge.
\newblock In \emph{Proceedings of ACL}.

\bibitem[{Qi et~al.(2018)Qi, Lin, Sun, Zhu, Xie, and Liu}]{qi2018cross}
Fanchao Qi, Yankai Lin, Maosong Sun, Hao Zhu, Ruobing Xie, and Zhiyuan Liu.
  2018.
\newblock Cross-lingual lexical sememe prediction.
\newblock In \emph{Proceedings of EMNLP}.

\bibitem[{Qi et~al.(2021{\natexlab{b}})Qi, Xie, Zang, Liu, and
  Sun}]{qi2021sememe}
Fanchao Qi, Ruobing Xie, Yuan Zang, Zhiyuan Liu, and Maosong Sun.
  2021{\natexlab{b}}.
\newblock Sememe knowledge computation: a review of recent advances in
  application and expansion of sememe knowledge bases.
\newblock \emph{Frontiers of Computer Science}, 15(5):1--11.

\bibitem[{Qi et~al.(2022)Qi, Yang, Yi, Cheng, Liu, and Sun}]{qi2022quoter}
Fanchao Qi, Yanhui Yang, Jing Yi, Zhili Cheng, Zhiyuan Liu, and Maosong Sun.
  2022.
\newblock Quoter: A benchmark of quote recommendation for writing.
\newblock In \emph{Proceedings of ACL}.

\bibitem[{Qi et~al.(2021{\natexlab{c}})Qi, Yao, Xu, Liu, and Sun}]{qi2021turn}
Fanchao Qi, Yuan Yao, Haoji Xu, Zhiyuan Liu, and Maosong Sun.
  2021{\natexlab{c}}.
\newblock Turn the combination lock: Learnable textual backdoor attacks via
  word substitution.
\newblock In \emph{Proceedings of ACL}.

\bibitem[{Qin et~al.(2020)Qin, Qi, Ouyang, Liu, Yang, Wang, Liu, and
  Sun}]{qin2020improving}
Yujia Qin, Fanchao Qi, Sicong Ouyang, Zhiyuan Liu, Cheng Yang, Yasheng Wang,
  Qun Liu, and Maosong Sun. 2020.
\newblock Improving sequence modeling ability of recurrent neural networks via
  sememes.
\newblock \emph{IEEE/ACM Transactions on Audio, Speech, and Language
  Processing}.

\bibitem[{Ridnik et~al.(2021)Ridnik, Ben-Baruch, Noy, and
  Zelnik-Manor}]{ridnik2021imagenet}
Tal Ridnik, Emanuel Ben-Baruch, Asaf Noy, and Lihi Zelnik-Manor. 2021.
\newblock Imagenet-21k pretraining for the masses.
\newblock In \emph{Proceedings of NeurIPS}.

\bibitem[{Sch{\"o}lkopf et~al.(1999)Sch{\"o}lkopf, Williamson, Smola,
  Shawe-Taylor, and Platt}]{scholkopf1999support}
Bernhard Sch{\"o}lkopf, Robert Williamson, Alex Smola, John Shawe-Taylor, and
  John Platt. 1999.
\newblock Support vector method for novelty detection.
\newblock In \emph{Proceedings of NIPS}.

\bibitem[{Vrande{\v{c}}i{\'c} and Kr{\"o}tzsch(2014)}]{vrandevcic2014wikidata}
Denny Vrande{\v{c}}i{\'c} and Markus Kr{\"o}tzsch. 2014.
\newblock Wikidata: a free collaborative knowledgebase.
\newblock \emph{Communications of the ACM}, 57(10):78--85.

\bibitem[{Vyas and Carpuat(2016)}]{vyas2016sparse}
Yogarshi Vyas and Marine Carpuat. 2016.
\newblock Sparse bilingual word representations for cross-lingual lexical
  entailment.
\newblock In \emph{Proceedings of NAACL-HLT}.

\bibitem[{Wierzbicka(1996)}]{wierzbicka1996semantics}
Anna Wierzbicka. 1996.
\newblock \emph{Semantics: Primes and universals: Primes and universals}.
\newblock Oxford University Press, UK.

\bibitem[{Wolf et~al.(2020)Wolf, Chaumond, Debut, Sanh, Delangue, Moi, Cistac,
  Funtowicz, Davison, Shleifer et~al.}]{wolf2020transformers}
Thomas Wolf, Julien Chaumond, Lysandre Debut, Victor Sanh, Clement Delangue,
  Anthony Moi, Pierric Cistac, Morgan Funtowicz, Joe Davison, Sam Shleifer,
  et~al. 2020.
\newblock Transformers: State-of-the-art natural language processing.
\newblock In \emph{Proceedings of EMNLP}.

\bibitem[{Xie et~al.(2017{\natexlab{a}})Xie, Liu, Luan, and Sun}]{xie2017image}
Ruobing Xie, Zhiyuan Liu, Huanbo Luan, and Maosong Sun. 2017{\natexlab{a}}.
\newblock Image-embodied knowledge representation learning.
\newblock In \emph{Proceedings of IJCAI}.

\bibitem[{Xie et~al.(2017{\natexlab{b}})Xie, Yuan, Liu, and
  Sun}]{xie2017lexical}
Ruobing Xie, Xingchi Yuan, Zhiyuan Liu, and Maosong Sun. 2017{\natexlab{b}}.
\newblock Lexical sememe prediction via word embeddings and matrix
  factorization.
\newblock In \emph{Proceedings of IJCAI}.

\bibitem[{Zang et~al.(2020)Zang, Qi, Yang, Liu, Zhang, Liu, and
  Sun}]{zang2020word}
Yuan Zang, Fanchao Qi, Chenghao Yang, Zhiyuan Liu, Meng Zhang, Qun Liu, and
  Maosong Sun. 2020.
\newblock Word-level textual adversarial attacking as combinatorial
  optimization.
\newblock In \emph{Proceedings of ACL}.

\bibitem[{Zhang et~al.(2020)Zhang, Qi, Liu, Wang, Liu, and
  Sun}]{zhang2020multi}
Lei Zhang, Fanchao Qi, Zhiyuan Liu, Yasheng Wang, Qun Liu, and Maosong Sun.
  2020.
\newblock Multi-channel reverse dictionary model.
\newblock In \emph{Proceedings of AAAI}.

\end{thebibliography}

\appendix

\end{CJK}
\end{document}